\documentclass[preprint,12pt]{elsarticle}
\usepackage{natbib}
\usepackage{hyperref}

\usepackage{amssymb}
\usepackage{amsmath}

\usepackage{booktabs}
\usepackage{multirow}
\usepackage{graphicx}
\usepackage{makecell}

\usepackage{float}
\usepackage{url}

\begin{document}

\begin{frontmatter}



\title{SAC$^2$-Net: Semantic Anchoring and Complementary-Consensus Fusion for Multimodal Micro-Expression Recognition}

\author[label1,label3,label4]{Xuepeng Zheng}
\ead{xpzheng@email.swu.edu.cn}
\author[label2,label3,label4]{Tong Chen\corref{mycorresponding}}
\cortext[mycorresponding]{Corresponding author}
\ead{c_tong@swu.edu.cn}
\author[label1,label3,label4]{Chaoping Gui}
\ead{gcp1017@126.com}
\author[label1,label3,label4]{Yingjuan Jia}
\ead{jyj151005@email.swu.edu.cn}
\author[label1,label3,label4]{Hanpu Wang}
\ead{wanghp568@gmail.com}
\author[label1,label3,label4]{Yuhao Shan}
\ead{shanyuhao@swu.edu.cn}

\affiliation[label1]{organization={College of Electronic and Information Engineering},
            addressline={Southwest University},
            city={Chongqing},
            postcode={400715},
            country={China}}

\affiliation[label2]{organization={Faculty of psychology},
			addressline={Southwest University},
        	city={Chongqing},
        	postcode={400715},
        	country={China}}
        	
\affiliation[label3]{organization={Chongqing Key Laboratory of Generic Technology and System of Service Robots},
			addressline={Southwest University},
			city={Chongqing},
			postcode={400715},
			country={China}}
	
\affiliation[label4]{organization={Institute of Legal Psychology and Intelligent Computing},
			addressline={Southwest University},
			city={Chongqing},
			postcode={400715},
			country={China}}

\begin{abstract}
Micro-expression recognition (MER) is challenging due to subtle facial movements, limited data, and the ambiguous relationship between Action Units (AUs) and emotion categories. Optical flow and motion magnification are two widely used representations for making subtle facial dynamics observable. However, many existing methods treat them as separate cues or fuse them without explicitly modeling their dual complementarity. Optical flow encodes displacement-level muscle motion, whereas motion magnification reveals appearance-level changes in facial texture and context. When both modalities are informative, their combination provides a more complete characterization of subtle facial dynamics; when one modality degrades, the other may still preserve discriminative evidence for compensation. This dual complementarity provides richer facial representations, but also introduces two key challenges for multimodal fusion: cross-modal heterogeneity and spatially varying modality reliability. To address these challenges, we propose SAC$^2$-Net, a Semantic Anchoring and Complementary-Consensus Network that first aligns heterogeneous visual representations with semantic anchors and then performs reliability-aware complementary fusion. Specifically, Semantic Anchoring Soft Alignment (SASA) converts activated AUs into textual prompts and uses hierarchical AU-aware soft labels to align motion-magnified and optical-flow representations while preserving semantic proximity among anatomically related samples. Based on the aligned representations, Complementary-Consensus Fusion (CCF) exchanges complementary motion and appearance cues, adaptively enhances unreliable local responses with trustworthy cross-modal evidence, and further encourages a shared spatial focus through consensus refinement. Extensive experiments on five MER benchmarks demonstrate that SAC$^2$-Net achieves state-of-the-art or highly competitive performance across coarse-grained, fine-grained, large-scale, and cross-dataset evaluation settings. Code is available at \url{https://github.com/pong213/SAC2-Net}.
\end{abstract}




\end{frontmatter}




\section{Introduction}
\label{sec:introduction}

Facial expressions are generally divided into macro-expressions (MaEs) and micro-expressions (MEs). MaEs are relatively obvious, last longer, and are often consciously displayed, whereas MEs are involuntary, brief, and subtle facial movements that occur when an individual attempts to conceal or suppress a genuine emotion~\cite{ekman2009telling}. Lasting fewer than 500 milliseconds and involving extremely weak muscle activations, MEs carry critical information for applications in clinical diagnosis, deception detection, and affective computing~\cite{yan2013casme,li2022deep}. MER aims to identify the emotional category conveyed by a detected ME segment, such as happiness, sadness, or surprise. Nevertheless, MER remains challenging because the visual signal is weak and often invisible to the naked eye, making it difficult for conventional deep models to learn discriminative representations from raw frames alone.

Early MER research primarily analyzed complete video sequences by extracting spatiotemporal descriptors over the entire clip to characterize facial dynamics~\cite{pfister2011recognising}. However, processing every frame is computationally expensive and introduces considerable temporal redundancy. Later studies found that the most discriminative facial changes are usually concentrated around several key frames, especially the apex frame, where an ME reaches its peak intensity. This motivated key-frame-based MER, in which the onset, apex, and offset frames correspond to the start, peak, and end of facial muscle movements, respectively. In particular, several studies have shown that using the apex frame, or compact onset--apex representations, can preserve emotion-relevant information while substantially reducing redundancy~\cite{liong2018less}.

Within this key-frame-based paradigm, raw frames are commonly transformed into representations that make subtle facial dynamics more explicit. According to the Facial Action Coding System (FACS)~\cite{ekman1978facial}, facial expressions can be decomposed into a set of Action Units (AUs), each corresponding to the contraction or relaxation of specific facial muscles. Optical flow estimated between the onset and apex frames is one of the most widely used representations for MER~\cite{liong2019ststnet,gan2019off}. It provides a motion-centric description of local facial displacement and highlights where AU-related movements occur, capturing dynamics that are difficult to perceive from individual frames. Another important way to make weak facial changes observable is motion magnification, which amplifies subtle appearance variations caused by micro-level muscle movements~\cite{wu2025am_mm_mer}. Unlike optical flow, motion-magnified images retain facial appearance context and reveal fine-grained local deformations in the image domain. In short, optical flow emphasizes displacement-level motion evidence, whereas motion magnification exposes appearance-level changes that may otherwise remain imperceptible. Both representations make weak facial signals more observable, yet each captures only a partial view of subtle facial dynamics.

\begin{figure}[t]
	\centering
	\includegraphics[width=.5\textwidth]{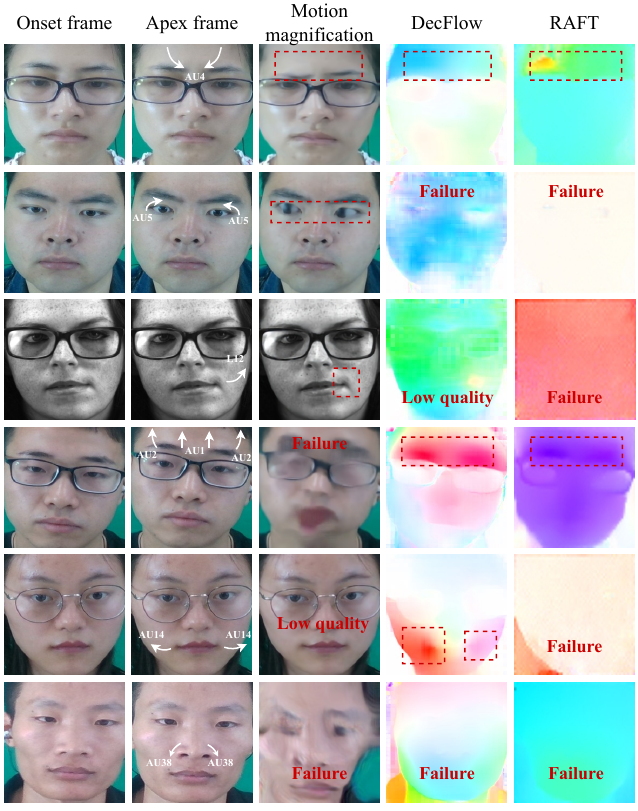}
	\caption{Examples of modality-specific success and failure cases. DecFlow~\cite{lu2024facialflownet} and RAFT~\cite{teed2020raft} are included to show that optical flow failure is not estimator-specific.}
	\label{fig:modal_failures_and_complementarity}
\end{figure}

Building on this distinction, our study starts from two observations regarding these two modalities. First, when both modalities are reliable, they provide complementary rather than redundant information. Optical flow contributes precise displacement cues, whereas motion magnification preserves richer appearance context, allowing the two modalities to jointly capture subtle facial dynamics more completely than either alone. Second, optical flow and motion magnification often fail in different ways. As illustrated in Fig.~\ref{fig:modal_failures_and_complementarity}, some samples exhibit distorted or artifact-contaminated magnified appearances, while their optical flow maps still preserve meaningful AU-related motion evidence. Conversely, some optical flow maps become nearly uniform or unreliable, whereas the corresponding magnified images still reveal discriminative facial changes. This asymmetric failure pattern indicates that their complementarity extends beyond information diversity to reliability. Therefore, an effective multimodal MER framework should simultaneously exploit complementary motion and appearance cues when both modalities are informative, while adaptively transferring trustworthy evidence from reliable regions of one modality to compensate for unreliable regions in the other.

However, exploiting this complementarity is nontrivial. First, optical flow and motion magnification have different representation forms and statistical properties, leading to cross-modal heterogeneity before fusion. If the two modalities are directly fused, the fusion module must simultaneously resolve semantic discrepancy, suppress modality-specific noise, and perform fine-grained emotion discrimination. Second, although MEs are closely related to AUs, AUs do not have a one-to-one correspondence with emotion categories. Samples with overlapping or anatomically related AU patterns may be semantically close even if their emotion labels differ. Treating such samples as unrelated negatives in contrastive learning can disturb the semantic structure of the representation space. Third, local modality quality varies across facial regions. Blind concatenation or generic attention fusion may either overlook complementary information or propagate unreliable evidence.

To address these issues, we propose SAC$^2$-Net (Semantic Anchoring and Complementary-Consensus Network), a multimodal MER framework designed to exploit the intrinsic complementarity and asymmetric reliability of motion magnification and optical flow. SAC$^2$-Net follows an align-first, fuse-later principle. Before fusion, we propose Semantic Anchoring Soft Alignment (SASA) to reduce cross-modal heterogeneity. Specifically, SASA converts activated AUs into natural-language prompts and uses them as stable semantic anchors for aligning motion-magnified and optical-flow representations. The AU-derived textual anchors provide explicit, dataset-agnostic, and fine-grained semantic guidance during training. More importantly, SASA constructs hierarchical AU-aware soft labels, so that samples with overlapping or anatomically related AU patterns remain close in the embedding space rather than being pushed apart as hard negatives. In this way, SASA imposes AU-aware semantic structure, allowing the downstream fusion module to focus on complementary information rather than large semantic discrepancies.

Based on the aligned representations, SAC$^2$-Net further introduces \\ Complementary-Consensus Fusion (CCF) to exploit local modality complementarity. Its Complementary Exchange Module (CEM) estimates spatial reliability maps for both modalities and allows each branch to exchange complementary motion and appearance cues from reliable regions of the other branch. When both modalities are informative, this interaction enriches feature representation by integrating displacement-level and appearance-level evidence. When one modality becomes unreliable, the same mechanism repairs degraded local responses using trustworthy information from the other modality. Its Consensus Refinement Module (CRM) then constructs a shared spatial attention reference, encouraging both modalities to focus on the same discriminative facial regions while preserving modality-specific content. Thus, CEM promotes complementary information exchange and reliability-aware feature repair, while CRM further stabilizes the fusion by encouraging spatial consensus.

The main contributions of this work are summarized as follows:
\begin{itemize}
	\item We identify the intrinsic information complementarity and asymmetric reliability between motion magnification and optical flow in MER, and propose SAC$^2$-Net, a multimodal framework that exploits complementary motion-appearance information.
	
	\item We propose SASA, a text-anchored soft alignment strategy that uses AU-derived prompts as stable semantic anchors and leverages hierarchical AU similarity to construct soft labels, thereby reducing cross-modal heterogeneity and preserving semantic proximity among related samples.
	
	\item We propose CCF, a multimodal fusion mechanism composed of CEM and CRM, which performs reliability-aware complementary exchange and consensus refinement to jointly exploit complementary information and repair unreliable local evidence.
	
	\item Experiments under various evaluation settings demonstrate the robustness and generality of SAC$^2$-Net. Ablation and qualitative analyses further demonstrate the effectiveness of the proposed semantic alignment and complementary-consensus fusion mechanisms.
	
\end{itemize}



\section{Related work}
\label{sec:related_work}

\subsection{Single-modal MER}
\label{sec:rw_single}

Optical flow is widely used in MER because it explicitly encodes the direction and magnitude of subtle facial muscle motion. Recent methods have improved optical-flow-based representation learning through contrastive learning, attention mechanisms, graph modeling, and Transformers. For example, SRMCL~\cite{bao2024srmcl} introduces self-expression reconstruction and prototype-based memory contrastive learning to improve generalization under limited MER data. HTNet~\cite{wang2024htnet} divides the face into key regions and applies hierarchical local-to-global self-attention on optical flow and optical strain. EMRNet~\cite{liu2025emrnet} employs channel-wise region-aware attention and distance-correlation constraints to enhance the diversity of optical-flow representations. MPFNet~\cite{ma2025mpfnet} combines multiple prior-learning strategies, while LTR3O~\cite{zhu2025ltr3o} reduces dependence on manually specified apex frames by ranking onset-occurring-offset representations. Although these methods refine motion modeling, optical flow remains vulnerable to weak motion, low texture, and noisy estimation. Moreover, since AUs do not map one-to-one to emotion categories, motion cues alone provide limited semantic evidence for fine-grained discrimination.

Motion magnification offers another way to enhance ME signals by amplifying subtle facial changes into more perceivable appearance variations. CMNet~\cite{wei2023cmnet} introduces contrastive magnification learning to capture intensity-related motion cues and model temporal intensity evolution. AM-MM-MER~\cite{wu2025am_mm_mer} first amplifies subtle facial motions using a Swin Transformer-based motion magnification network, and then performs recognition through Transformer modeling of selected facial landmarks. These methods make subtle expressions more observable and preserve facial appearance context that optical flow may discard. However, motion magnification is also sensitive to preprocessing quality. Excessive or inaccurate magnification may introduce artifacts or distortions, thereby misrepresenting the true expression.

In summary, optical flow and motion magnification provide complementary displacement- and appearance-level views of facial dynamics, while also exhibiting modality-specific degradation. This motivates their joint use.

\subsection{Multimodal MER}
\label{sec:rw_multi}

Multimodal MER methods aim to improve robustness by combining heterogeneous visual cues. Li et al.~\cite{li2026of_mag_cl} adopt regional weighted fusion with supervised contrastive learning, and MMTNet~\cite{wang2025mmtnet} fuses dynamic images and optical strain using multi-scale Transformers and cross-modal contrastive alignment. MFDAN~\cite{cai2024mfdan} further uses optical flow to guide attention over magnified image features. Although these methods improve visual representation learning, most of them align or fuse visual modalities directly, without explicitly modeling AU-level semantic relations or the spatially varying reliability of each modality.

AU-guided semantic learning provides another important direction. AU-GACN~\cite{xie2020augacn} introduces AU supervision and graph attention to model AU-emotion relationships. More recently, MER-CLIP~\cite{liu2025merclip} aligns visual ME dynamics with AU-guided natural-language prompts through CLIP-style contrastive learning. However, its hard positive-negative formulation is less suitable for samples with overlapping AU patterns, especially among similar negative emotions such as anger, fear, disgust, and sadness. Existing AU-guided methods also rarely consider hierarchical anatomical relations, such as bilateral-unilateral AU correspondence.



\section{Method}

\begin{figure}[t]
	\centering
	\includegraphics[width=\textwidth]{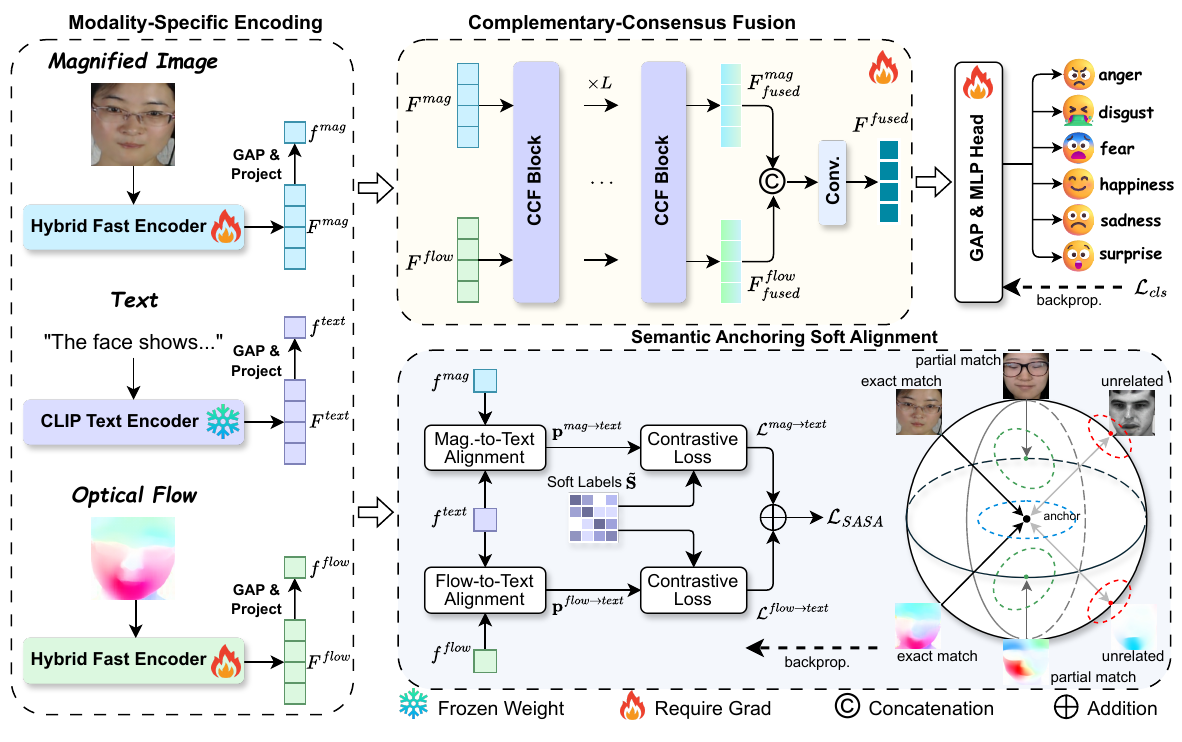}
	\caption{Overall architecture of the proposed SAC$^2$-Net.}
	\label{fig:overall_architecture}
\end{figure}

The overall architecture of SAC$^2$-Net is shown in Fig.~\ref{fig:overall_architecture}. During training, AU-derived textual prompts serve as semantic anchors to guide the representation learning of the two visual modalities. While the textual modality provides stable semantic grounding, motion magnification and optical flow provide complementary appearance- and displacement-level evidence with spatially varying reliability. Accordingly, SASA aligns both visual modalities to the AU-derived text space using anatomically informed soft relational supervision. The aligned features are then fed into CCF, which performs complementary information exchange, reliability-aware refinement, and spatial consensus learning for final recognition.

\subsection{Modality-specific encoding}
\label{sec:feature_extraction}

\subsubsection{Hierarchical visual encoding}
\label{sec:visual_encoder}

MER requires both local AU-related motion cues and long-range dependencies among facial regions. To capture these two types of information, we adopt a hybrid visual backbone termed the Hybrid Fast Encoder (HFE), which combines convolutional token mixing in shallow stages with self-attention in the deepest stage.

HFE follows a four-stage hierarchical MetaFormer-style architecture~\cite{vasu2023fastvit}. Each stage is composed of repeated residual blocks:
\begin{align}
	Y' &= X + \mathrm{TokenMixer}(\mathrm{Derf}(X)), \\
	Y &= Y' + \mathrm{ConvFFN}(\mathrm{Derf}(Y')), 
\end{align}
where $\mathrm{Derf}(\cdot)$ denotes the Dynamic erf~\cite{chen2026derf}, $\mathrm{TokenMixer}(\cdot)$ performs spatial information aggregation, and $\mathrm{ConvFFN}(\cdot)$ is a ConvNeXt-style feed-forward block~\cite{liu2022convnet}. Stages 1--3 use depthwise separable convolution to capture local motion patterns, while Stage 4 uses multi-head self-attention to model global facial dependencies. The detailed configuration is provided in Supplementary material.

Let $\mathcal{E}_{v}^{mag}(\cdot)$ and $\mathcal{E}_{v}^{flow}(\cdot)$ denote the visual encoders for motion magnification and optical flow, respectively. They share the same HFE architecture but do not share parameters because the two modalities have different appearance statistics and noise patterns. Given the motion-magnified image $X^{mag}\in\mathbb{R}^{3\times H\times W}$ and the optical-flow image $X^{flow}\in\mathbb{R}^{3\times H\times W}$, the spatial feature maps are extracted as
\begin{align}
	F^{mag}  &= \mathcal{E}_{v}^{mag}(X^{mag})  \in \mathbb{R}^{D\times H'\times W'}, \\
	F^{flow} &= \mathcal{E}_{v}^{flow}(X^{flow}) \in \mathbb{R}^{D\times H'\times W'}.
\end{align}
These spatial features are retained for subsequent complementary fusion. For semantic alignment, we further apply global average pooling (GAP) and modality-specific projection heads:
\begin{align}
	f^{mag}  &= g_{mag}\big(\mathrm{GAP}(F^{mag})\big) \in \mathbb{R}^{D}, \\
	f^{flow} &= g_{flow}\big(\mathrm{GAP}(F^{flow})\big) \in \mathbb{R}^{D},
\end{align}
where $g_{mag}(\cdot)$ and $g_{flow}(\cdot)$ are learnable projection heads.

\subsubsection{Textual prompt generation and encoding}
\label{sec:text_encoder}

\noindent\textbf{Stochastic AU-to-text prompting.}
To avoid overfitting to fixed sentence patterns, we design a stochastic prompting strategy. For each AU, we build a template set containing descriptions from complementary perspectives, such as anatomical movement and visible appearance. We also maintain multiple sentence prefixes to diversify the global prompt context.

Let $\mathcal{A}={a_1,a_2,\ldots,a_K}$ denote the activated AU set of a sample. For each AU $a_k$, one description is randomly sampled from its template set $\mathcal{D}(a_k)$ and composed under a randomly sampled context prefix $\mathcal{T}_{base}(\cdot)$:
\begin{equation}
	T = \mathcal{T}_{base}\!\big(\mathcal{D}(a_1) \textcircled{c}  \mathcal{D}(a_2)\textcircled{c} \cdots \textcircled{c} \mathcal{D}(a_K)\big),
\end{equation}
where $\textcircled{c}$ denotes concatenation. For example, for AU labels $\{4,7\}$, one possible prompt is: \emph{The facial expression is characterized by the eyebrows being pulled downward and inward, together with tightened and narrowed eyelids.} This strategy acts as semantic data augmentation, encouraging the model to focus on AU-related meaning rather than fixed syntax. The complete prompt library is released with the code.

\noindent\textbf{Textual encoding.}
The generated prompt is encoded by the text encoder of CLIP~\cite{radford2021learning} and projected into the shared alignment space:
\begin{equation}
	f^{text} = g_t\big(\mathcal{E}_t(T)\big) \in \mathbb{R}^{D},
\end{equation}
where $\mathcal{E}_t(\cdot)$ denotes the CLIP text encoder and $g_t(\cdot)$ is a learnable projection head. We freeze $\mathcal{E}_t$ during training to preserve its pre-trained semantic knowledge and reduce overfitting on small MER datasets.


\subsection{Semantic anchoring soft alignment}
\label{sec:sasa}

Standard contrastive learning typically assumes binary sample relations, i.e., each pair is either a positive or a negative. Such a formulation is overly restrictive for MER, where different samples often share partial facial-action semantics. For instance, two samples may correspond to different expression categories yet still exhibit overlapping AU activations. Treating such semantically related pairs as hard negatives may distort the learned embedding space. To address this issue, we propose SASA, a text-anchored contrastive learning strategy that replaces binary supervision with AU-conditioned soft relational targets.

\subsubsection{AU-based soft label generation}
\label{sec:soft_labels}

\noindent\textbf{Pairwise AU similarity.}
In FACS, some AUs may appear either bilaterally or unilaterally. For example, AU4 may appear as bilateral activation affecting both eyebrows, or as L4/R4 indicating left-only or right-only activation. Standard set-overlap measures ignore this anatomical relation. We therefore define an AU-family mapping $\rho(\cdot)$ that removes laterality while retaining the AU identity:
\begin{equation}
	\rho(4)=\rho(L4)=\rho(R4)=4.
\end{equation}
Since each sample contains at most one label from the same AU family, cross-set matching at the family level is unique.

For two AU labels $a$ and $b$, their anatomically informed pairwise similarity is defined as
\begin{equation}
	\sigma(a,b)=
	\begin{cases}
		1, & a=b,\\
		\alpha, & \rho(a)=\rho(b), \text{one bilateral and one unilateral},\\
		\beta, & \rho(a)=\rho(b), \text{opposite unilateral sides},\\
		0, & \text{otherwise},
	\end{cases}
	\label{eq:pairwise_sigma}
\end{equation}
where $1>\alpha>\beta>0$. Here, $\alpha$ and $\beta$ model the anatomical similarity between bilateral--unilateral and opposite-side unilateral activations, respectively. In our experiments, $\alpha=0.7$ and $\beta=0.6$. Their effects are analyzed in Supplementary material.

\begin{figure}[t]
	\centering
	\includegraphics[width=.7\textwidth]{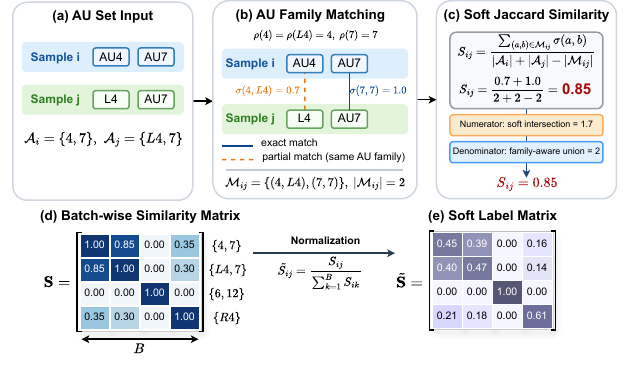}
	\caption{Example of AU-based soft label generation in SASA.}
	\label{fig:soft_label_gen}
\end{figure}

\noindent\textbf{Soft Jaccard similarity.}
Let $\mathcal{A}_i$ and $\mathcal{A}_j$ denote the AU sets of samples $i$ and $j$. The matched AU-family pairs are defined as
\begin{equation}
	\mathcal{M}_{ij}=\{(a,b)\mid a\in\mathcal{A}_i,\; b\in\mathcal{A}_j,\; \rho(a)=\rho(b)\}.
\end{equation}
The AU-level soft Jaccard similarity is then computed as
\begin{equation}
	S_{ij}=
	\frac{
		\sum_{(a,b)\in\mathcal{M}_{ij}} \sigma(a,b)
	}{
		|\mathcal{A}_i|+|\mathcal{A}_j|-|\mathcal{M}_{ij}|
	}.
	\label{eq:soft_jaccard}
\end{equation}
This formulation preserves the union-overlap property of Jaccard similarity while incorporating partial anatomical correspondence within the same AU family. To obtain soft contrastive targets, we normalize the batch-wise similarity matrix:
\begin{equation}
	\tilde{S}_{ij}=\frac{S_{ij}}{\sum_{k=1}^{B} S_{ik}},
	\label{eq:row_norm}
\end{equation}
where $B$ is the batch size. The normalized matrix $\tilde{\mathbf{S}}\in\mathbb{R}^{B\times B}$ is used as the relational target for SASA. Fig.~\ref{fig:soft_label_gen} illustrates the soft label generation process.

\subsubsection{Text-anchored contrastive learning}
\label{sec:text_anchor}

Because motion magnification and optical flow are generated by different preprocessing pipelines, they may contain different noise patterns and modality-specific distortions. Directly aligning them may therefore force unreliable visual evidence into the shared representation. SASA instead adopts an asymmetric text-anchored alignment strategy, where both visual modalities are aligned to the AU-derived text branch, which serves as a stable semantic reference.

Given the visual embedding $f_i^{v}$ and textual embedding $f_j^{text}$, where $v\in\{mag,flow\}$, the visual-to-text alignment probability is computed as
\begin{align}
	p_{ij}^{v\rightarrow text}
	=
	\frac{
		\exp\left(\mathrm{sim}(f_i^{v},f_j^{text})/\tau\right)
	}{
		\sum_{k=1}^{B}
		\exp\left(\mathrm{sim}(f_i^{v},f_k^{text})/\tau\right)
	}, \\
	\label{eq:alignment_prob}
	\mathrm{sim}(f^{v},f^{text}) = \frac{
		f^{v} \cdot {(f^{text})}^\top
	}{\|f^{v}\| \cdot \|f^{text}\|} \nonumber,
\end{align}
where $\tau$ is the temperature parameter. The alignment loss is defined as the soft-target cross-entropy between the predicted alignment distribution and the AU-conditioned relational target:
\begin{equation}
	\mathcal{L}^{v\rightarrow text}
	=
	-\frac{1}{B}
	\sum_{i=1}^{B}
	\sum_{j=1}^{B}
	\tilde{S}_{ij}\log p_{ij}^{v\rightarrow text}.
	\label{eq:soft_ce}
\end{equation}
The overall SASA loss is
\begin{equation}
	\mathcal{L}_{\mathrm{SASA}}
	=
	\mathcal{L}^{mag\rightarrow text}
	+
	\mathcal{L}^{flow\rightarrow text}.
	\label{eq:sasa_total}
\end{equation}

By preserving graded AU-level relations, SASA structures the visual embedding space before fusion. This reduces cross-modal semantic discrepancy and allows the subsequent CCF module to exploit modality complementarity under spatially varying reliability.


\subsection{Complementary-consensus fusion}
\label{sec:ccf}

Although SASA reduces cross-modal semantic discrepancy, the two modalities still encode complementary motion and appearance information while exhibiting spatially varying reliability. To exploit both properties, we propose Complementary-Consensus Fusion (CCF), a two-stage exchange-then-agree fusion block. CEM enables complementary information exchange between modalities while adaptively enhancing unreliable local responses, and CRM subsequently encourages both modalities to establish a shared spatial reference without suppressing modality-specific characteristics.

\begin{figure}[t]
	\centering
	\includegraphics[width=\textwidth]{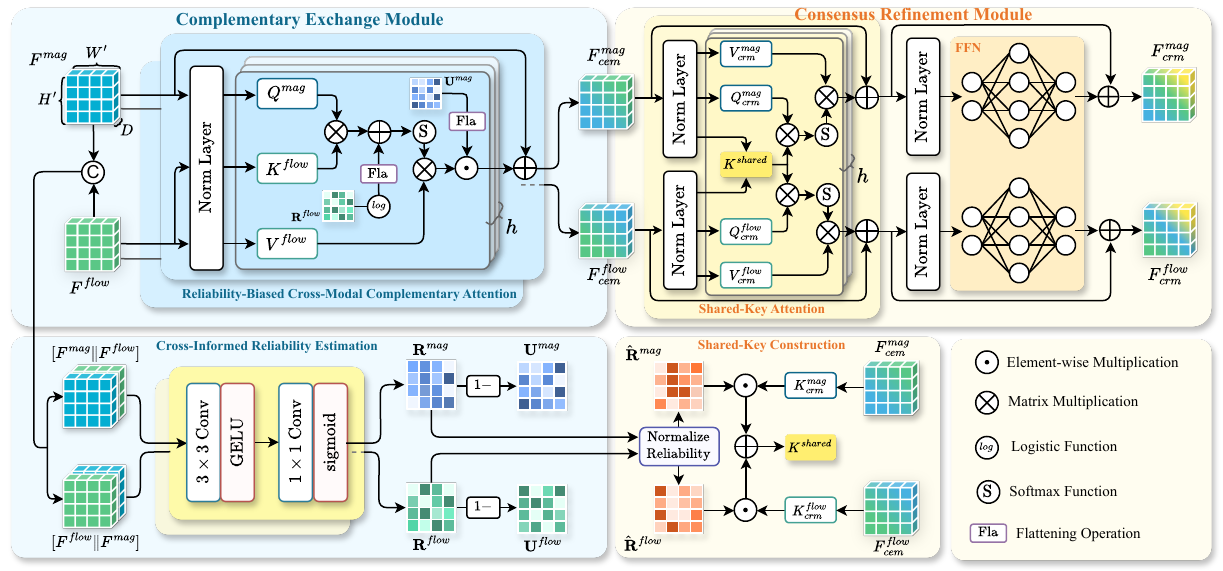}
	\caption{Architecture of the proposed CCF block. CEM performs reliability-aware complementary exchange between semantically aligned motion-magnified and optical-flow features, followed by CRM for spatial consensus refinement. For attention computation in CEM and CRM, the input feature maps are serialized from $\mathbb{R}^{B\times D\times H'\times W'}$ to $\mathbb{R}^{B\times N\times D}$, where $N=H'W'$.}
	\label{fig:ccf_architecture}
\end{figure}

\subsubsection{Complementary exchange module}
\label{sec:cem}

CEM aims to exploit complementary motion and appearance information while repairing unreliable local responses. To this end, it jointly models the reliability of each modality and the spatial locations where complementary evidence should be exchanged. Let $v\in\{mag,flow\}$ and let $\bar{v}$ denote the modality complementary to $v$. Given aligned feature maps $F^{mag},F^{flow}\in\mathbb{R}^{D\times H'\times W'}$, the cross-informed reliability map of modality $v$ is estimated as
\begin{equation}
	\mathbf{R}^{v}
	=
	\sigma\!\Big(
	\mathrm{Conv}_{1}^{v}\big(
	\mathrm{GELU}\left(
	\mathrm{Conv}_{3}^{v}([F^{v}\|F^{\bar{v}}])
	\right)
	\big)
	\Big),
	\label{eq:cem_reliability}
\end{equation}
where $\mathbf{R}^{v}\in\mathbb{R}^{1\times H'\times W'}$, $\sigma(\cdot)$ denotes the sigmoid function, and $\|$ denotes channel-wise concatenation. The estimator uses both streams so that the reliability of one modality is judged with reference to the other. The unreliability map is defined as $\mathbf{U}^{v}=\mathbf{1}-\mathbf{R}^{v}$.

These reliability maps are learned jointly with the recognition objective and are further constrained by a dedicated regularization loss introduced in Sec.~\ref{sec:optimization}.

After flattening the spatial dimensions into sequences of length $N=H'W'$, CEM performs reliability-biased cross-modal attention. For modality $v$, the complementary feature borrowed from $\bar{v}$ is computed as
\begin{equation}
	F^{v\leftarrow \bar{v}}
	=
	\mathrm{softmax}
	\left(
	\frac{Q^{v}(K^{\bar{v}})^\top}{\sqrt{d_k}}
	+
	\log(\mathbf{R}^{\bar{v}}_{seq}+\epsilon)
	\right)
	V^{\bar{v}},
	\label{eq:cem_attention}
\end{equation}
where $Q^{v}$ is projected from $F^{v}$, $K^{\bar{v}}$ and $V^{\bar{v}}$ are projected from $F^{\bar{v}}$, and $\mathbf{R}^{\bar{v}}_{seq}$ is the flattened reliability map of the source modality. The log-reliability term acts as an attention prior, guiding the target modality to retrieve information from trustworthy source regions.

The complementary feature is injected through an unreliability-gated residual update:
\begin{equation}
	F^{v}_{cem}
	=
	F^{v}
	+
	\mathbf{U}^{v}\odot F^{v\leftarrow \bar{v}},
	\label{eq:cem_update}
\end{equation}
where $\odot$ denotes element-wise multiplication with spatial broadcasting.

\subsubsection{Consensus refinement module}
\label{sec:crm}

After CEM, the two modalities should agree on the most discriminative facial regions while preserving their distinct representations. CRM is therefore introduced to establish spatial consensus. Instead of directly merging the two streams, CRM encourages consensus at the level of attention coordinates while retaining modality-specific values.

The reliability maps are first normalized as
\begin{equation}
	\hat{\mathbf{R}}^{v}
	=
	\frac{\mathbf{R}^{v}}
	{\mathbf{R}^{mag}+\mathbf{R}^{flow}+\epsilon}.
	\label{eq:normalized_reliability}
\end{equation}
A reliability-gated shared key is then constructed from the exchanged features:
\begin{equation}
	K^{shared}
	=
	\hat{\mathbf{R}}^{mag}\odot K^{mag}_{crm}
	+
	\hat{\mathbf{R}}^{flow}\odot K^{flow}_{crm},
	\label{eq:shared_key}
\end{equation}
where $K^{mag}_{crm}$ and $K^{flow}_{crm}$ are key projections from the two exchanged streams. The shared key provides a reliability-weighted spatial reference: regions dominated by one reliable modality contribute more strongly, while regions where both modalities are informative contribute jointly.

Each modality then computes attention using its own query and value but shares the same key:
\begin{equation}
	\Delta F^{v}_{crm}
	=
	\mathrm{softmax}
	\left(
	\frac{Q^{v}_{crm}(K^{shared})^\top}{\sqrt{d_k}}
	\right)
	V^{v}_{crm}.
	\label{eq:crm_attention}
\end{equation}
This design separates where to attend from what to retrieve: the shared key encourages both modalities to attend to consistent facial regions, whereas modality-specific values preserve their distinct representation capacity.

The CRM output is obtained by residual refinement:
\begin{align}
	F^{v}_{crm}
	&=
	F^{v}_{cem}
	+
	\Delta F^{v}_{crm}, \\
	F^{v}_{crm}
	&\leftarrow
	F^{v}_{crm}
	+
	\mathrm{FFN}(F^{v}_{crm}).
	\label{eq:crm_update}
\end{align}
After stacking $L$ CCF blocks, the final fused magnification and flow features are concatenated and mixed by a $1\times1$ convolution:
\begin{equation}
	F^{fused}
	=
	\mathrm{Conv}_1
	\left(
	[F^{mag}_{fused}\|F^{flow}_{fused}]
	\right)
	\in\mathbb{R}^{D\times H'\times W'}.
	\label{eq:final_fused}
\end{equation}


\subsection{Optimization objective}
\label{sec:optimization}

The overall objective consists of classification supervision, semantic anchoring, and reliability regularization:
\begin{equation}
	\mathcal{L}_{total}
	=
	\lambda_{cls}(e)\mathcal{L}_{cls}
	+
	\lambda_{SASA}(e)\mathcal{L}_{SASA}
	+
	\lambda_{rel}\mathcal{L}_{rel},
	\label{eq:total_loss}
\end{equation}
where $\mathcal{L}_{SASA}$ is defined in Eq.~\eqref{eq:sasa_total}, and $e$ denotes the training epoch.

\noindent\textbf{Classification loss.}
To reduce the effect of class imbalance in MER datasets, we adopt a class-balanced focal loss:
\begin{equation}
	\mathcal{L}_{cls}
	=
	-\frac{1}{B}
	\sum_{i=1}^{B}
	w_{y_i}(1-p_{i,y_i})^{\gamma}\log p_{i,y_i},
	\label{eq:focal_loss}
\end{equation}
where $p_{i,y_i}$ is the predicted probability of the ground-truth class $y_i$, $\gamma$ is the focusing parameter, and $w_{y_i}$ is the class weight. Following the effective-number formulation~\cite{cui2019class}, the weight of class $c$ is computed as
\begin{equation}
	w_c
	=
	\frac{1-\beta}{1-\beta^{n_c}},
	\label{eq:effective_num}
\end{equation}
where $n_c$ is the number of training samples in class $c$, and $\beta=(N-1)/N$ with $N$ denoting the total number of training samples.

\noindent\textbf{Reliability regularization.}
Since the reliability maps in CEM guide complementary exchange and consensus construction, they are encouraged to be spatially smooth. We regularize them using a total-variation-style loss:
\begin{equation}
	\mathcal{L}_{rel}
	=
	\sum_{v\in{mag,flow}}
	\left(
	\|\nabla_x \mathbf{R}^{v}\|_1
	+
	\|\nabla_y \mathbf{R}^{v}\|_1
	\right),
	\label{eq:rel_loss}
\end{equation}
where $\nabla_x(\cdot)$ and $\nabla_y(\cdot)$ denote horizontal and vertical finite differences.

\noindent\textbf{Dynamic loss scheduling.}
Semantic anchoring is emphasized in the early stage to organize the visual embedding space, while classification supervision is gradually strengthened for final emotion recognition. To keep the total contribution of these two objectives stable, we adopt the following cosine schedule:
\begin{align}
	\lambda_{SASA}(e) &= \frac{\Lambda}{2}
	\left(
	1+\cos\!\left(\frac{\pi e}{E-1}\right)
	\right), \\
	\lambda_{cls}(e) &= \Lambda-\lambda_{SASA}(e),
	\label{eq}
\end{align}
where $E$ is the total number of training epochs and $\Lambda=2$. Thus, $\lambda_{SASA}(e)$ decreases from $2$ to $0$, while $\lambda_{cls}(e)$ increases from $0$ to $2$. The reliability regularization weight is fixed to $\lambda_{rel}=0.1$.



\section{Experiments}
\label{sec:experiments}


\subsection{Datasets and evaluation protocols}
\label{sec:datasets}

We evaluate SAC$^2$-Net on five publicly available spontaneous ME datasets.

\noindent\textbf{CASME~II}~\cite{yan2014casme} contains 247 ME samples collected from 26 subjects at 200 fps. Following common practice, experiments on this dataset are conducted under the 5-class setting, including happiness, disgust, repression, surprise, and others.

\noindent\textbf{SAMM}~\cite{davison2016samm} consists of 159 samples from 32 subjects recorded at 200 fps. Although the dataset provides seven emotion categories, we follow the commonly adopted 5-class setting (happiness, contempt, surprise, anger, and others).

\noindent\textbf{SMIC}~\cite{li2013spontaneous} comprises 164 samples from 16 subjects captured at 100 fps and annotated with three coarse-grained emotion labels, namely positive, negative, and surprise. 
Since SMIC does not provide AU annotations, we additionally annotated the activated AUs following the FACS protocol.

\noindent\textbf{MEGC2019-CD}~\cite{see2019megc} was introduced by the ME Grand Challenge 2019 (MEGC2019) to enable cross-dataset evaluation under a unified label space. Following the official protocol, SMIC, CASME~II, and SAMM are combined into a composite benchmark, where all samples are mapped into three coarse-grained emotion categories: positive, negative, and surprise.

\noindent\textbf{CAS(ME)$^3$}~\cite{li2022cas} is a large-scale dataset containing 1,109 samples from 247 subjects with high-resolution depth information. In this work, we use Part A of CAS(ME)$^3$, which contains 860 samples from 100 subjects. Experiments are conducted under both the 4-class setting (positive, negative, surprise, and others) and the 7-class setting (happiness, disgust, surprise, fear, anger, sadness, and others).

\noindent\textbf{DFME}~\cite{zhao2023dfme} is one of the largest publicly available ME datasets containing 7,526 samples from 656 subjects across seven emotion categories, including happiness, disgust, contempt, surprise, fear, anger, and sadness. In our experiments, we use its publicly available subset, which is partitioned into a training set (1,856 samples), Test-A set (474 samples), and Test-B set (299 samples).

For evaluation, we adopt leave-one-subject-out (LOSO) cross-validation on all datasets except DFME. For DFME, we follow the protocol in~\cite{zhao2024dynamic} and conduct experiments using its predefined training and testing split. To further assess cross-dataset generalization, we conduct two transfer settings by using CASME~II and SAMM as the training set, respectively, and SMIC as the test set. Following common practice in MER, we report Accuracy (ACC), F1-score (F1), unweighted F1-score (UF1), and unweighted average recall (UAR), which are defined as
\begin{align}
	ACC &= \frac{\sum_{c=1}^{C} TP_c}{\sum_{c=1}^{C} N_c}, \\
	F1 &= \sum_{c=1}^{C}\frac{N_c}{N}\frac{2TP_c}{2TP_c + FP_c + FN_c}, \\
	UF1 &= \frac{1}{C}\sum_{c=1}^{C}\frac{2TP_c}{2TP_c + FP_c + FN_c}, \\
	UAR &= \frac{1}{C}\sum_{c=1}^{C}\frac{TP_c}{N_c},
\end{align}
where $C$ denotes the number of classes, $N_c$ is the number of samples in the $c$-th class, and $TP_c$, $FP_c$, and $FN_c$ denote the numbers of true positives, false positives, and false negatives for the $c$-th class, respectively.


\subsection{Implementation details}
\label{sec:implementation}

In all experiments, facial regions are first aligned and cropped using MediaPipe~\cite{lugaresi2019mediapipe}, and then resized to $224 \times 224$. The onset and apex frames of each sample are used to construct the two visual modalities. Specifically, the motion-magnified image is generated by the learning-based axial motion magnification method in~\cite{byung2024learning} with the axial magnification factors set to $5$, while the optical-flow image is estimated using DecFlow~\cite{lu2024facialflownet}.

To initialize the visual encoder, we first perform SASA-based contrastive pretraining on the external macro-expression dataset Extended Cohn-Kanade (CK+)~\cite{lucey2010extended}. This external pretraining helps the HFE learn semantically structured facial motion representations before downstream MER training. During this stage, only the HFE visual encoders and SASA-related projection layers are optimized, while the subsequent fusion module and classifier are excluded. The pretrained weights are then transferred to downstream MER datasets for task-specific training.

SAC$^2$-Net is optimized using AdamW~\cite{loshchilov2017decoupled} with an initial learning rate of $5 \times 10^{-4}$ and a weight decay of $5 \times 10^{-2}$. A cosine annealing schedule is employed after a $3$-epoch linear warm-up. The model is trained for $200$ epochs with a batch size of $64$. To stabilize fine-tuning, the visual encoders are updated with a reduced learning rate equal to $0.2\times$ the base learning rate, while the fusion module and classifier are trained using the full base rate. The number of CCF blocks $L$ is set to 3. Unless otherwise specified, all experiments are repeated 3 times with different random seeds, and the average results are reported as the final performance. All experiments are implemented in PyTorch and conducted on two NVIDIA RTX A100 GPUs.


\subsection{Comparison with state-of-the-art}
\label{sec:sota}

\begin{table}[h]
	\centering
	\caption{Comparison results on SMIC, CASME II, SAMM, and MEGC2019-CD under the 3-class setting. The best results are highlighted in bold, and the second-best results are underlined. Methods marked with $^{*}$ are trained using multiple modalities.
	}
	\label{tab:megc2019_cd}
	\resizebox{\linewidth}{!}{
		\begin{tabular}{lccccccccc}
			\toprule
			\multirow{2}{*}{Methods} 
			& \multirow{2}{*}{Pub-Year}
			& \multicolumn{2}{c}{SMIC} 
			& \multicolumn{2}{c}{CASME II} 
			& \multicolumn{2}{c}{SAMM} 
			& \multicolumn{2}{c}{MEGC2019-CD} \\
			\cmidrule(lr){3-4} \cmidrule(lr){5-6} \cmidrule(lr){7-8} \cmidrule(lr){9-10}
			& & UF1 & UAR & UF1 & UAR & UF1 & UAR & UF1 & UAR \\
			\midrule
			$\mu$-BERT~\cite{nguyen2023micron}        & CVPR-23  & \textbf{85.50} & \underline{83.84} & 90.34 & 89.14 & 83.86 & 84.75 & 89.03 & 88.42 \\
			FRL-DGT~\cite{zhai2023feature}           & CVPR-23  & 74.30 & 74.90 & 91.90 & 90.30 & 77.20 & 75.80 & 81.20 & 81.10 \\
			SRMCL~\cite{bao2024srmcl}             & TAC-24   & 79.46 & 80.53 & 96.35 & 96.49 & 84.70 & \underline{88.66} & 86.30 & 88.30 \\
			HTNet~\cite{wang2024htnet}             & NC-24    & 80.49 & 79.05 & 95.32 & 95.16 & 81.31 & 81.24 & 86.03 & 84.75 \\
			MFDAN$^{*}$~\cite{cai2024mfdan}       & TCSVT-24 & 68.15 & 70.43 & 91.34 & 93.26 & 78.71 & 81.96 & 84.53 & 86.88 \\
			EMRNet~\cite{liu2025emrnet}            & AIR-25   & 65.09 & 65.96 & 90.74 & 89.95 & 67.82 & 68.97 & 74.68 & 75.46 \\
			MPFNet~\cite{ma2025mpfnet}            & TAC-25   & 80.60 & 80.90 & 91.10 & 92.30 & 79.50 & 83.90 & 84.00 & 84.60 \\
			Micro\_NesT~\cite{he2025micro_nest}       & ESWA-25  & 78.86 & 78.16 & \underline{97.22} & \underline{96.58} & 86.64 & 87.31 & 87.22 & 86.57 \\
			EDMDBN~\cite{ma2025edmdbn}            & PRL-25   & 79.48 & 80.85 & 94.84 & 96.19 & 83.36 & 86.61 & 88.21 & \underline{89.33} \\
			MMTNet$^{*}$~\cite{wang2025mmtnet}      & JVCIR-25 & 81.03 & 80.72 & 87.62 & 86.90 & 77.89 & 75.52 & 83.21 & 82.33 \\
			LTR3O~\cite{zhu2025ltr3o}             & TAC-25   & 83.36 & 82.98 & 95.78 & 94.87 & 89.12 & 85.26 & \underline{89.31} & 88.19 \\
			MOL~\cite{shao2025mol}               & TPAMI-25 & 81.00 & 72.34 & 90.08 & 89.92 & \textbf{89.72} & \textbf{89.00} & 87.79 & 85.42 \\
			SAC$^2$-Net            & Ours     & \underline{84.22} & \textbf{84.83} & \textbf{97.33} & \textbf{96.88} & \underline{89.17} & 84.32 & \textbf{89.73} & \textbf{89.35} \\
			\bottomrule
		\end{tabular}
	}
\end{table}


\noindent\textbf{Coarse-grained dataset evaluation.}
As shown in Table~\ref{tab:megc2019_cd}, SAC$^2$-Net achieves the best overall performance on the CASME~II and MEGC2019-CD datasets. Table~\ref{tab:casme3} further shows it obtains the highest UF1 and UAR on the 4-class setting of CAS(ME)$^3$. We attribute this consistent performance to the text-anchored alignment. Since AU descriptions are dataset-agnostic, they provide a stable semantic scaffold that generalizes across heterogeneous recording conditions. In addition, SASA's soft labels preserve hierarchical AU similarity, preventing semantically related classes from being overly separated during training and thereby improving coarse-grained discrimination.

Nevertheless, SAC$^2$-Net does not achieve the best result on the SAMM subset of the composite benchmark. This may be due to the limited scale and severe class imbalance of SAMM, where the positive and surprise categories are under-represented. Such scarcity weakens the AU-similarity-based relational supervision used by SASA, especially for minority classes. Since UF1 and UAR equally weight each class, errors on these categories have a disproportionate effect, narrowing the advantage of SAC$^2$-Net on this subset.

\begin{table}[h]
	\centering
	\caption{Comparison results on CASME II and SAMM under the 5-class setting. -- indicates that the corresponding paper did not report results under this setting.}
	\label{tab:casme2_samm}
	\resizebox{.6\linewidth}{!}{
		\begin{tabular}{lccccc}
			\toprule
			\multirow{2}{*}{Methods} 
			& \multirow{2}{*}{Pub-Year}
			& \multicolumn{2}{c}{CASME II}
			& \multicolumn{2}{c}{SAMM} \\
			\cmidrule(lr){3-4} \cmidrule(lr){5-6}
			& & ACC & UF1 & ACC & UF1 \\
			\midrule
			$\mu$-BERT~\cite{nguyen2023micron}        & CVPR-23  & 83.48 & 85.53 & \textbf{84.75} & \textbf{83.86} \\
			FRL-DGT~\cite{zhai2023feature}           & CVPR-23  & 75.70 & 74.80 & -- & -- \\
			SRMCL~\cite{bao2024srmcl}             & TAC-24   & 83.20 & 82.86 & 74.63 & 65.99 \\
			MPFNet~\cite{ma2025mpfnet}            & TAC-25   & 83.10 & 83.30 & 71.90 & 71.80 \\
			Micro\_NesT~\cite{he2025micro_nest}       & ESWA-25  & 77.93 & 77.20 & 76.69 & 74.78 \\
			EDMDBN~\cite{ma2025edmdbn}            & PRL-25   & \underline{88.26} & 85.91 & 81.58 & 75.73 \\
			MER-CLIP$^{*}$~\cite{liu2025merclip}    & TAC-25   & 82.33 & 83.78 & 77.21 & 74.14 \\
			MMTNet$^{*}$~\cite{wang2025mmtnet}      & JVCIR-25 & 80.08 & 80.07 & 75.00 & 67.36 \\
			LTR3O~\cite{zhu2025ltr3o}             & TAC-25   & 81.78 & 79.05 & 80.15 & 75.74 \\
			GAMDSS~\cite{liu2026gamdss}            & TAC-26   & 87.50 & \underline{86.17} & 82.84 & 81.47 \\
			SAC$^2$-Net            & Ours     & \textbf{88.34} & \textbf{87.87} & \underline{83.38} & \underline{82.09} \\
			\bottomrule
		\end{tabular}
	}
\end{table}

\begin{table}[h]
	\centering
	\caption{Comparison results on CAS(ME)$^3$ under the 4-class and 7-class settings. $^{\dagger}$ indicates that the relevant results are reproduced using official code of the corresponding method.
	}
	\label{tab:casme3}
	\resizebox{.7\linewidth}{!}{
		\begin{tabular}{lccccc}
			\toprule
			\multirow{2}{*}{Methods} 
			& \multirow{2}{*}{Pub-Year}
			& \multicolumn{2}{c}{4-class}
			& \multicolumn{2}{c}{7-class} \\
			\cmidrule(lr){3-4} \cmidrule(lr){5-6}
			& & UF1 & UAR & UF1 & UAR \\
			\midrule
			AlexNet~\cite{li2022cas}            & TPAMI-22 & 29.15 & 29.10 & 17.59 & 18.01 \\
			$\mu$-BERT~\cite{nguyen2023micron}         & CVPR-23  & 47.18 & 49.13 & 32.64 & 32.54 \\
			HTNet$^{\dagger}$~\cite{wang2024htnet}  & NC-24    & 49.35 & 52.34 & 37.56 & 37.69 \\
			SFAMNet~\cite{liong2024sfamnet}            & NC-24    & 44.62 & 47.97 & 23.65 & 23.73 \\
			PC-GCN~\cite{zhang2025facial}             & TAC-25   & 47.64 & 53.66 & 35.64 & 41.59 \\
			MER-CLIP$^{*}$~\cite{liu2025merclip}     & TAC-25   & 65.44 & 62.42 & 49.97 & 50.14 \\
			GAMDSS~\cite{liu2026gamdss}             & TAC-26   & \underline{70.75} & \underline{76.03} & \underline{53.29} & \textbf{62.73} \\
			SAC$^2$-Net             & Ours     & \textbf{72.33} & \textbf{77.34} & \textbf{56.99} & \underline{58.28} \\
			\bottomrule
		\end{tabular}
	}
\end{table}

\noindent\textbf{Fine-grained dataset evaluation.}
Table~\ref{tab:casme2_samm} reports results under the more challenging fine-grained classification setting. On CASME~II, SAC$^2$-Net achieves the best ACC and UF1, while maintaining competitive performance on SAMM. On CAS(ME)$^3$ with the 7-class setting (Table~\ref{tab:casme3}), SAC$^2$-Net obtains the best UF1 and competitive UAR. Notably, its gain over the strongest baseline increases from $+1.58\%$ in the 4-class setting to $+3.70\%$ in the 7-class setting, suggesting that the advantage of SAC$^2$-Net becomes more pronounced as class granularity increases. Fine-grained MER requires distinguishing emotions with similar AU patterns, such as anger and disgust, both of which involve AU4. In such cases, CEM helps exploit the modality that better captures subtle local differences, while CRM guides both modalities toward consistent discriminative facial regions instead of relying on inconsistent cues.

Consistent with the coarse-grained evaluation, SAC$^2$-Net achieves competitive but non-leading performance on SAMM. Its limited scale becomes more restrictive under fine-grained classification, where splitting the dataset into more emotion categories further reduces the number of samples per class. Such sparsity makes it difficult to form reliable AU-similarity neighborhoods and learn stable class-specific modality interactions. In contrast, the larger CAS(ME)$^3$ dataset provides denser fine-grained supervision, allowing the advantages of SASA, CEM, and CRM to become more evident as class granularity increases.

\begin{table}[h]
	\centering
	\caption{Comparison results on DFME under the 7-class setting. The results of Wang et al. and He et al. are reported by~\cite{liu2025merclip}.}
	\label{tab:dfme}
	\resizebox{.7\linewidth}{!}{
		\begin{tabular}{lcccc}
			\toprule
			\multirow{2}{*}{Methods} 
			& \multirow{2}{*}{Test Set}
			& \multicolumn{3}{c}{7-class} \\
			\cmidrule(lr){3-5}
			& & UF1 & UAR & ACC \\
			\midrule
			HTNet$^{\dagger}$~\cite{wang2024htnet}       & \multirow{5}{*}{Set A} & 42.35 & 42.83 & 49.14 \\
			Wang et al.~\cite{zhao2024dynamic}    &                        & 40.67 & 40.74 & 46.41 \\
			He et al.~\cite{zhao2024dynamic}      &                        & 41.23 & 42.10 & 48.73 \\
			MER-CLIP$^{*}$~\cite{liu2025merclip}          &                        & \underline{50.24} & \underline{51.15} & \underline{56.96} \\
			SAC$^2$-Net                  &                        & \textbf{52.73} & \textbf{52.90} & \textbf{58.65} \\
			\midrule
			HTNet$^{\dagger}$~\cite{wang2024htnet}       & \multirow{5}{*}{Set B} & 36.36 & 37.71 & 39.25 \\
			Wang et al.~\cite{zhao2024dynamic}    &                        & 35.34 & 36.61 & 38.13 \\
			He et al.~\cite{zhao2024dynamic}      &                        & 40.16 & 40.08 & 41.47 \\
			MER-CLIP$^{*}$~\cite{liu2025merclip}          &                        & \textbf{51.28} & \textbf{51.20} & \textbf{52.50} \\
			SAC$^2$-Net                  &                        & \underline{50.68} & \underline{50.57} & \underline{51.50} \\
			\bottomrule
		\end{tabular}
	}
\end{table}

\noindent\textbf{Large-scale dataset evaluation.}
Table~\ref{tab:dfme} reports the 7-class results on the DFME benchmark. On Test set~A, SAC$^2$-Net achieves the best performance across all metrics. On Test set~B, SAC$^2$-Net ranks second, trailing MER-CLIP by modest margins. Notably, both methods outperform the remaining baselines by over $10\%$, suggesting that cross-modal semantic guidance is critical for large-scale MER. The complementary exchange mechanism plays a significant role at this scale, since the larger sample pool introduces greater variability in the quality of motion magnification and optical flow across samples, making reliability-aware cross-modal interaction especially beneficial for stable prediction.

\begin{table}[!t]
	\centering
	\caption{Cross-dataset comparison results from CASME II and SAMM to SMIC under the 3-class setting. Following the protocol in~\cite{xie2020augacn}, models are trained on CASME~II or SAMM and tested on SMIC to evaluate cross-dataset generalization.
	}
	\label{tab:cross_dataset}
	\begin{tabular}{lccccc}
		\toprule
		\multirow{2}{*}{Methods} 
		& \multirow{2}{*}{Pub-Year}
		& \multicolumn{2}{c}{CASME II $\rightarrow$ SMIC}
		& \multicolumn{2}{c}{SAMM $\rightarrow$ SMIC} \\
		\cmidrule(lr){3-4} \cmidrule(lr){5-6}
		& & Acc & F1 & Acc & F1 \\
		\midrule
		STCNN~\cite{reddy2019stcnn}       & IJCNN-19   & 31.40 & 19.00 & 32.50 & 19.00 \\
		CapsuleNet~\cite{van2019capsulenet}  & FG-19      & 32.20 & 15.20 & 32.40 & 17.90 \\
		MER-GCN~\cite{lo2020mer}     & MIPR-20    & 36.70 & 27.20 & 36.10 & 17.80 \\
		AU-GACN~\cite{xie2020augacn}     & ACM MM-20  & 34.40 & 31.90 & \underline{45.10} & 30.90 \\
		MOL~\cite{shao2025mol}         & TPAMI-25   & \underline{47.13} & \underline{43.91} & 44.58 & \underline{32.32} \\
		SAC$^2$-Net      & Ours       & \textbf{56.46} & \textbf{56.60} & \textbf{52.17} & \textbf{45.41} \\
		\bottomrule
	\end{tabular}
\end{table}

\noindent\textbf{Cross-dataset evaluation.}
As shown in Table~\ref{tab:cross_dataset}, SAC$^2$-Net substantially outperforms all baselines in both transfer directions, demonstrating strong cross-dataset generalization. The SASA provides dataset-agnostic semantic guidance through AU-based text anchoring and thus encourages the model to learn transferable representations rather than dataset-specific patterns.


\subsection{Ablation study}
\label{sec:ablation}

\begin{table}[t]
	\centering
	\caption{Ablation study of the main components in SAC$^2$-Net.}
	\label{tab:ablation_module}
	\begin{tabular}{ccccc}
		\toprule
		\multirow{2}{*}{SASA} 
		& \multirow{2}{*}{CEM} 
		& \multirow{2}{*}{CRM} 
		& \multicolumn{2}{c}{CAS(ME)$^3$ (7-class)} \\
		\cmidrule(lr){4-5}
		& & & UF1 & UAR \\
		\midrule
		&         &         & 45.23 & 46.37 \\
		\checkmark &         &         & 49.07 & 50.13 \\
		\checkmark & \checkmark &         & 54.12 & 55.47 \\
		\checkmark & \checkmark & \checkmark & \textbf{56.99} & \textbf{58.28} \\
		\bottomrule
	\end{tabular}
\end{table}

\begin{table}[t]
	\centering
	\caption{Ablation study of different SASA label-generation strategies.}
	\label{tab:ablation_sasa}
	\begin{tabular}{ccccc}
		\toprule
		\multirow{2}{*}{\makecell{Hard\\labels}}
		& \multirow{2}{*}{\makecell{Standard\\Jaccard}}
		& \multirow{2}{*}{\makecell{Soft\\Jaccard}}
		& \multicolumn{2}{c}{CAS(ME)$^3$ (7-class)} \\
		\cmidrule(lr){4-5}
		& & & UF1 & UAR \\
		\midrule
		\checkmark &            &            & 53.42 & 54.80 \\
		& \checkmark &            & 55.21 & 56.74 \\
		&            & \checkmark & \textbf{56.99} & \textbf{58.28} \\
		\bottomrule
	\end{tabular}
\end{table}

\begin{table}[t]
	\centering
	\caption{Ablation study of the reliability estimation strategy in CEM and the shared-key construction strategy in CRM. Self-only and cross-informed reliability estimation are compared to evaluate the effect of cross-modal reliability modeling, while average and reliability-gated shared keys are compared to assess the importance of reliability-aware consensus construction.
	}
	\label{tab:ablation_ccf}
	\resizebox{\linewidth}{!}{
		\begin{tabular}{cccccc}
			\toprule
			\multicolumn{2}{c}{Reliability Estimation} 
			& \multicolumn{2}{c}{Shared-key} 
			& \multicolumn{2}{c}{CAS(ME)$^3$ (7-class)} \\
			\cmidrule(lr){1-2} \cmidrule(lr){3-4} \cmidrule(lr){5-6}
			Self-only & Cross-informed & Average & Reliability-gated & UF1 & UAR \\
			\midrule
			\checkmark &            & \checkmark &            & 52.35 & 53.41 \\
			& \checkmark & \checkmark &            & 55.04 & 56.60 \\
			\checkmark &            &            & \checkmark & 53.76 & 54.87 \\
			& \checkmark &            & \checkmark & \textbf{56.99} & \textbf{58.28} \\
			\bottomrule
		\end{tabular}
	}
\end{table}

\begin{table}[t]
	\centering
	\caption{Ablation study of the loss-weighting schedule on CAS(ME)$^3$ and CASME II.}
	\label{tab:ablation_schedule}
	\begin{tabular}{lcccc}
		\toprule
		\multirow{2}{*}{Schedule}
		& \multicolumn{2}{c}{CAS(ME)$^3$ (7-class)}
		& \multicolumn{2}{c}{CASME II (5-class)} \\
		\cmidrule(lr){2-3} \cmidrule(lr){4-5}
		& UF1 & UAR & UF1 & UAR \\
		\midrule
		Fixed   & 54.17 & 55.38 & 85.62 & 85.20 \\
		Dynamic & \textbf{56.99} & \textbf{58.28} & \textbf{87.87} & \textbf{88.48} \\
		\bottomrule
	\end{tabular}
\end{table}

\begin{table}[t]
	\centering
	\caption{Ablation study of the number of CCF blocks on different benchmarks.}
	\label{tab:ablation_ccf_blocks}
	\begin{tabular}{llccccc}
		\toprule
		\multirow{2}{*}{Benchmark} & \multirow{2}{*}{Metric}
		& \multicolumn{5}{c}{Number of CCF blocks} \\
		\cmidrule(lr){3-7}
		& & 1 & 2 & 3 & 4 & 5 \\
		\midrule
		\multirow{2}{*}{\begin{tabular}[c]{@{}c@{}}CASME II\\(5-class)\end{tabular}}
		& UF1 & 81.85 & 80.14 & \textbf{87.87} & 85.33 & 83.11 \\
		& UAR & 82.26 & 81.58 & \textbf{88.48} & 86.50 & 83.95 \\
		\midrule
		\multirow{2}{*}{\begin{tabular}[c]{@{}c@{}}SAMM\\(5-class)\end{tabular}}
		& UF1 & 77.68 & 76.32 & \textbf{82.09} & 77.31 & 73.82 \\
		& UAR & 76.49 & 75.64 & \textbf{82.36} & 76.95 & 72.69 \\
		\midrule
		\multirow{2}{*}{\begin{tabular}[c]{@{}c@{}}CAS(ME)$^3$\\(7-class)\end{tabular}}
		& UF1 & 51.10 & 50.74 & \textbf{56.99} & 54.60 & 53.21 \\
		& UAR & 52.35 & 50.80 & \textbf{58.28} & 56.24 & 55.79 \\
		\midrule
		\multirow{2}{*}{MEGC2019-CD}
		& UF1 & 82.61 & 87.31 & \textbf{89.73} & 86.42 & 87.46 \\
		& UAR & 83.92 & 88.26 & \textbf{89.35} & 86.07 & 87.29 \\
		\bottomrule
	\end{tabular}
\end{table}

Unless otherwise specified, all ablations are conducted on CAS(ME)$^3$ under the 7-class setting.

\noindent\textbf{Module-level ablations.}
Table~\ref{tab:ablation_module} isolates the contribution of each major component in SAC$^2$-Net. The baseline, which uses only the visual encoders followed by feature concatenation and classification, achieves $45.23\%$ UF1 and $46.37\%$ UAR. Adding SASA alone brings a gain of $3.84\%$ UF1, showing that semantic anchoring already provides a stronger representation space for MER. Introducing CEM further increases the improvement to $8.89\%$ UF1, indicating that reliability-aware complementary exchange effectively enhances cross-modal interaction. Incorporating CRM yields the full SAC$^2$-Net, reaching $56.99\%$ UF1 and $58.28\%$ UAR. Overall, the complete model improves the baseline by $11.76\%$ UF1 and $11.91\%$ UAR, demonstrating that SASA, CEM, and CRM contribute progressively and complementarily.

\noindent\textbf{Effect of SASA design.}
Table~\ref{tab:ablation_sasa} compares three variants of the alignment strategy, namely hard labels, standard Jaccard-based soft labels, and the proposed soft Jaccard labels. Replacing hard labels with soft labels brings clear gains, indicating that soft relational supervision is more suitable than binary positive/negative for MER, where samples often share partial AU semantics. Further replacing standard Jaccard with the proposed soft Jaccard yields the best performance, showing that encoding the anatomical structure of AUs, rather than only their co-occurrence, is essential for the alignment to reflect genuine semantic proximity.

\noindent\textbf{Effect of CCF design.}
Table~\ref{tab:ablation_ccf} analyzes two key design choices in CCF: the reliability estimation strategy in CEM and the shared-key construction strategy in CRM. Cross-informed reliability estimation contributes an average gain of $2.96\%$ UF1, confirming that reliability should be estimated with reference to both modalities rather than from each modality alone. Additionally, the reliability-gated shared key brings an  average gain of $1.68\%$ UF1. The best result is obtained when both designs are combined, showing that reliability information should be propagated throughout the fusion pipeline rather than introduced at only one stage.

\noindent\textbf{Effect of loss scheduling.}
Table~\ref{tab:ablation_schedule} compares fixed and dynamic loss weighting on CAS(ME)$^3$ (7-class) and CASME~II (5-class). The dynamic schedule consistently outperforms the fixed setting ($\lambda_{cls}=\lambda_{SASA}=1.0$) on both datasets. This supports our design intuition that semantic alignment should dominate the early stage of training to structure the embedding space, while classification should be emphasized later to optimize the final recognition objective.

\begin{figure}[t]
	\centering
	\includegraphics[width=.7\textwidth]{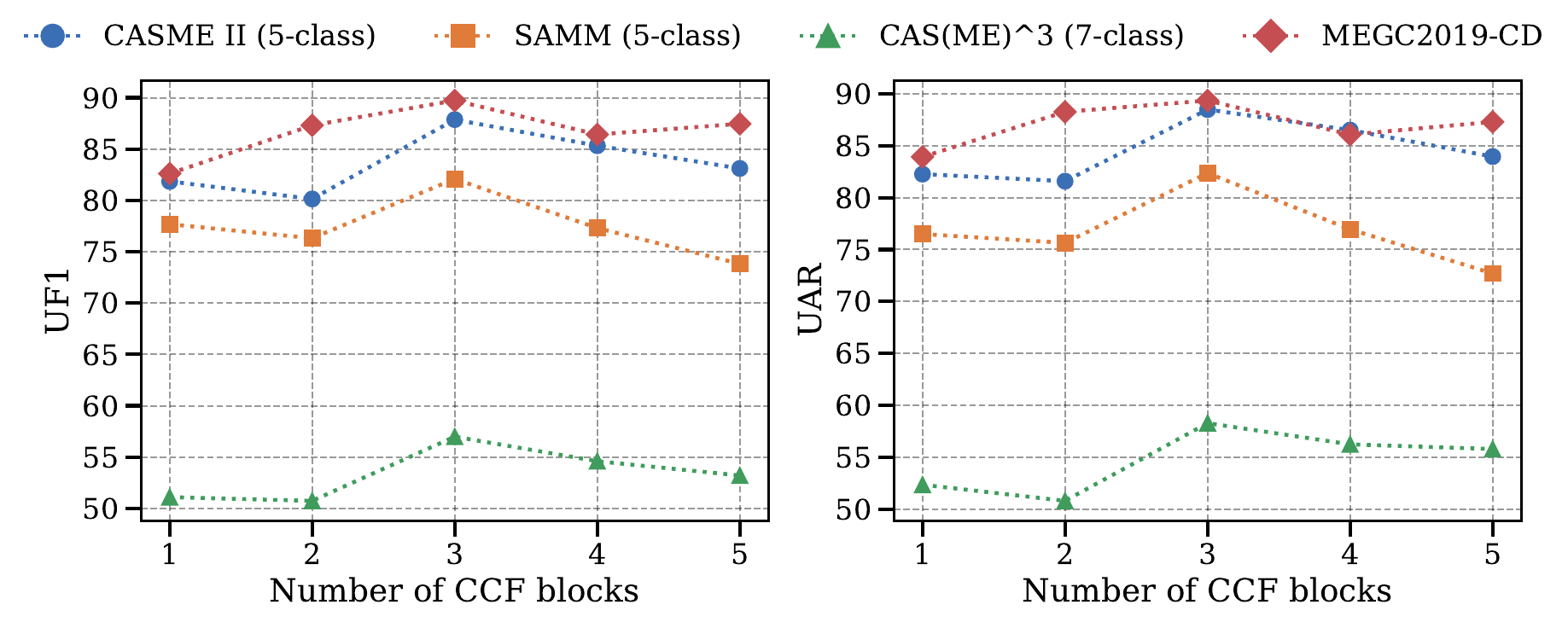}
	\caption{UF1 and UAR performance across four benchmarks with varying numbers of CCF blocks}
	\label{fig:ccf_depth}
\end{figure}

\noindent\textbf{Effect of the number of CCF blocks.}
To investigate the impact of CCF depth,  we vary the number of stacked CCF blocks across four datasets with different scales and label granularities. As shown in Table~\ref{tab:ablation_ccf_blocks} and illustrated in Fig.~\ref{fig:ccf_depth}, with fewer blocks, the fusion depth is insufficient to fully exploit complementary and consensus information. In contrast, stacking more than three blocks leads to consistent degradation, likely due to redundant feature refinement and overfitting on small-scale ME datasets (e.g., the SAMM dataset). The consistent results across four heterogeneous datasets suggest that setting $L=3$ provides a good balance between fusion capacity and generalization, rather than being an artifact of a particular dataset.

\subsection{Qualitative analysis}
\label{sec:qualitative}

\begin{figure}[t]
	\centering
	\includegraphics[width=\textwidth]{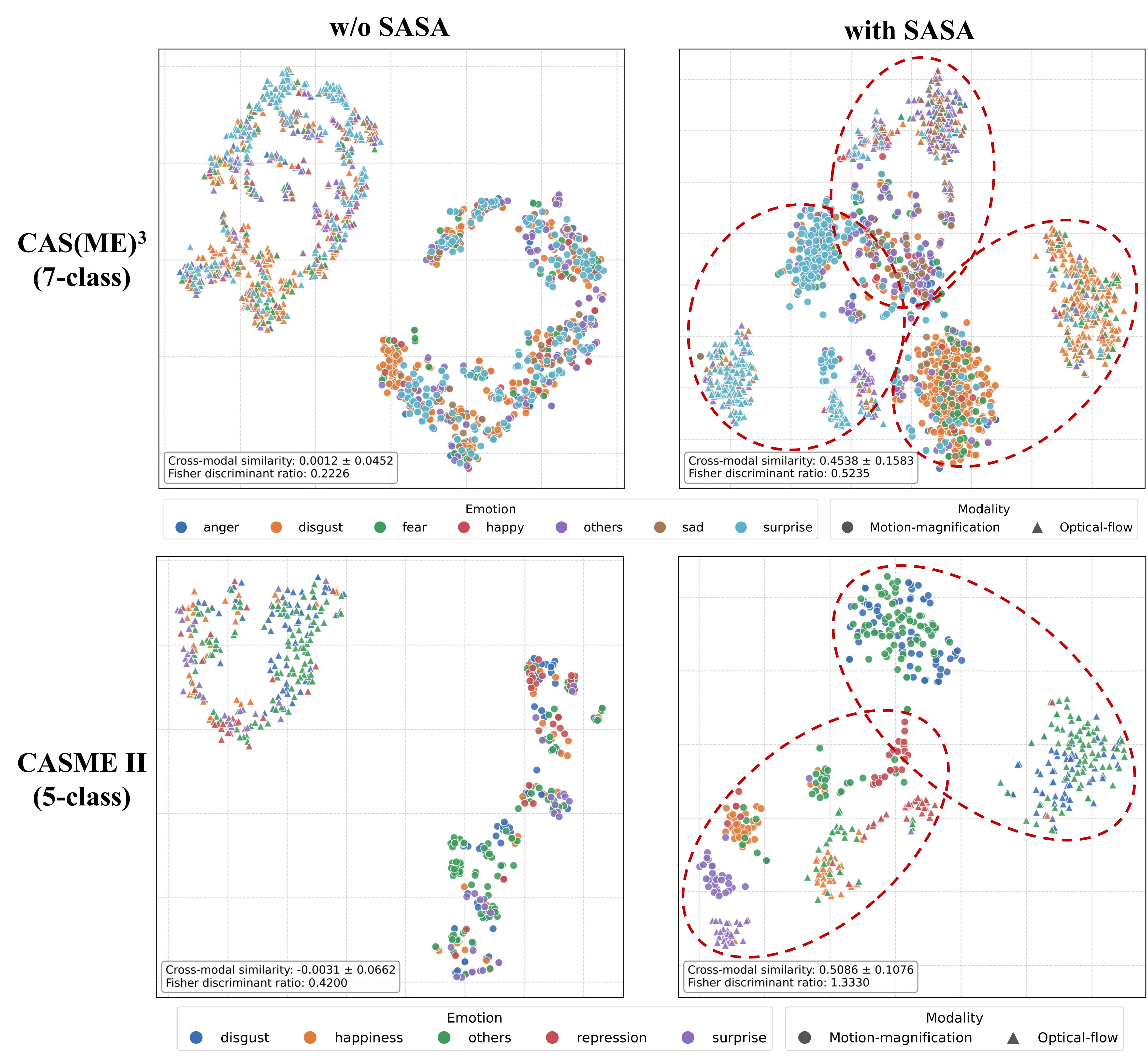}
	\caption{Training dynamics of semantic alignment on CAS(ME)$^3$ and CASME~II. The t-SNE snapshots show the evolution of motion-magnified and optical-flow embeddings at representative epochs. The curve subplot reports five epoch-wise metrics: mag-text and flow-text cosine similarities measure the alignment between each visual modality and AU-derived text anchors; mag-flow cosine similarity measures cross-modal consistency between the two visual modalities; FDR measures the semantic clustering of embedding space, computed as the ratio of inter-class center distance to intra-class dispersion.
	}
	\label{fig:tsne_sasa}
\end{figure}

\begin{figure}[t]
	\centering
	\includegraphics[width=.8\textwidth]{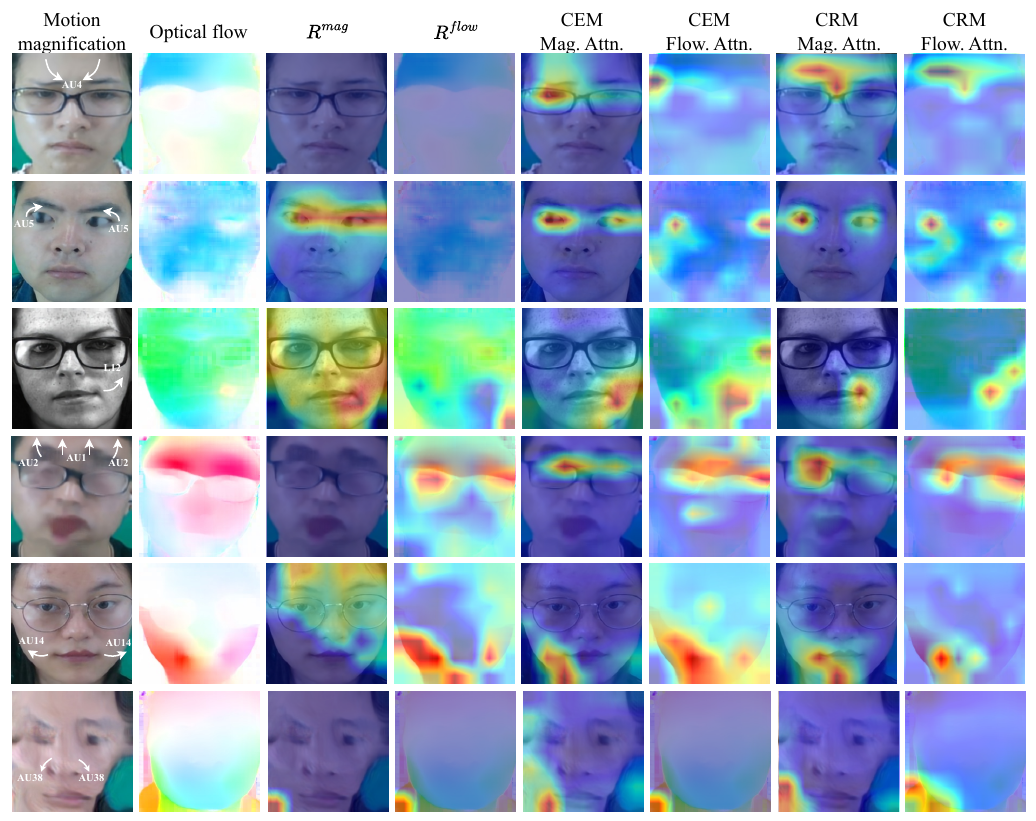}
	\caption{Visualization of reliability maps and attention responses in CCF. All reliability maps and attention maps are extracted from the last CCF block, since it is closest to the final prediction and best reflects the model's final complementary-consensus behavior. The reliability maps represent feature confidence rather than AU saliency; therefore, in clean samples they may appear relatively uniform instead of peaking only at activated AU regions.}
	\label{fig:cem_crm_vis}
\end{figure}

To better understand the behavior of SAC$^2$-Net, we provide qualitative analysis from three aspects: representation structuring by SASA, reliability-aware complementary exchange in CEM, and consensus refinement in CRM.

\noindent\textbf{Training dynamics of semantic alignment.}
To further examine how SASA influences representation learning throughout training, we visualize the projected embeddings of the motion-magnified and optical-flow branches on CAS(ME)$^3$ and CASME~II using t-SNE at representative epochs, and report the corresponding epoch-wise quantitative curves in Fig.~\ref{fig:tsne_sasa}. At the initial stage, the two visual modalities exhibit a clear modality gap and weak class-level structure. As training proceeds, especially in the early epochs where $\lambda_{\mathrm{SASA}}$ is large, the mag-text and flow-text cosine similarities~\cite{radford2021learning} increase rapidly, indicating that both visual branches are progressively aligned with the AU-derived semantic anchors. Meanwhile, the mag-flow cosine similarity also increases, showing that SASA indirectly improves cross-modal consistency by pulling both visual modalities toward the shared text-guided semantic space. The Fisher discriminant ratios (FDR)~\cite{fisher1936use} of both motion-magnified and optical-flow embeddings also rise during the early training stage, suggesting that SASA not only reduces inter-modal discrepancy but also improves class-level organization in the projected representation space.

The t-SNE snapshots provide a consistent visual interpretation of this process. From epoch 0 to epoch 50, samples from the two visual modalities become more closely aligned, and the class structure becomes clearer, reflecting the semantic organizing effect of SASA. Around epoch 100, as the SASA weight decreases and classification supervision becomes more dominant, the representation space becomes more discriminative. At epoch 200, although $\lambda_{\mathrm{SASA}}$ becomes small or approaches zero, the representation structure remains relatively consistent with that at epoch 100. This indicates that the semantic structure learned under SASA is largely preserved during later recognition-oriented optimization. Overall, these results demonstrate that SASA acts as an early semantic organizer: it first guides the two heterogeneous visual modalities into a more consistent and semantically structured embedding space, thereby providing a stronger foundation for subsequent multimodal fusion.

\noindent\textbf{Visualization of reliability-aware complementary exchange and consensus refinement.}
Fig.~\ref{fig:cem_crm_vis} visualizes the reliability maps and attention responses of CEM and CRM on six representative samples. In the first sample, where both modalities provide reliable evidence, CEM encourages complementary interaction between displacement-level motion and appearance-level context, producing concentrated responses around AU-related regions. CRM further aligns both branches toward a shared discriminative facial region. The second and third samples show cases where motion magnification remains informative but optical flow is degraded, whereas the fourth and fifth exhibit the opposite situation. In these asymmetric cases, CEM selectively enhances the degraded modality with complementary information from the more reliable one, and CRM further refines the shared spatial focus. By contrast, in the final sample, both modalities fail to preserve reliable AU-related evidence, causing CEM and CRM to attend to semantically irrelevant regions. These results demonstrate that CCF effectively exploits modality complementarity while adaptively handling spatially varying reliability, but remains challenged when both modalities are severely degraded.



\section{Conclusion}
\label{sec:conclusion}

In this paper, we presented a multimodal MER framework motivated by the intrinsic complementarity and asymmetric reliability of optical flow and motion magnification. SASA reduces cross-modal heterogeneity by aligning the two modalities with AU-derived semantic anchors, while CCF enables complementary information exchange, reliability-aware feature refinement, and shared spatial consensus. Extensive experiments across multiple evaluation settings demonstrate the robustness and generality of the proposed framework.

Despite these promising results, SAC$^2$-Net still has several limitations. First, when both motion magnification and optical flow fail to preserve reliable AU-related evidence, complementary fusion becomes less effective.. Second, SASA relies on AU annotations during training, making it sensitive to inaccurate or noisy labels. Third, the current framework is built on onset--apex representations and may not fully exploit the temporal evolution of micro-expression sequences. Future work will investigate more robust semantic supervision and sequence-level modeling for MER.


\section*{Acknowledgments}

This research is partly supported by Science and Technology Innovation Key R\&D Program of Chongqing (CSTB2023TIAD-STX0037)









\bibliographystyle{unsrtnat}
\bibliography{cas-refs}

\end{document}